\documentclass{article}
\PassOptionsToPackage{sort&compress,numbers,square}{natbib}
\usepackage[preprint]{neurips_2026}
\usepackage[utf8]{inputenc} % allow utf-8 input
\usepackage[T1]{fontenc}    % use 8-bit T1 fonts
\usepackage{hyperref}       % hyperlinks
\usepackage{url}            % simple URL typesetting
\usepackage{booktabs}       % professional-quality tables
\usepackage{amsfonts}       % blackboard math symbols
\usepackage{nicefrac}       % compact symbols for 1/2, etc.
\usepackage{microtype}      % microtypography
\usepackage[svgnames]{xcolor}         % colors

\usepackage[mode=buildnew]{standalone}
\usepackage{wrapfig}

\usepackage{amsmath, amssymb, amsthm, mathtools}
\usepackage[english]{babel}

\usepackage[inline]{enumitem}

\newtheorem{prop}{Proposition}
\newtheorem*{remark}{Remark}
\title{Temporal Rank in Recurrent Neural Networks}

\author{
    Leo Dobson \\
    School of Computer Science \\
    The University of Sheffield \\
    Sheffield, S1 4DP \\
    \texttt{ldobson4@sheffield.ac.uk} \\
    \And
    Anton Ragni \\
    School of Computer Science \\
    The University of Sheffield \\
    Sheffield, S1 4DP \\
    \texttt{a.ragni@sheffield.ac.uk} \\
}

\begin{document}

\maketitle

\begin{abstract}
    Recurrent Neural Networks (RNNs) are a natural framework for temporal modelling, yet are often dismissed due to their sequential, non-parallelisable computation and difficulty in capturing long-range dependencies, typically attributed to vanishing and exploding gradients. We identify a distinct and under-explored limitation: an information bottleneck induced by rank-constrained hidden state dynamics, which limits the number of independent input directions that can be preserved over time, as reflected in the rank of the input-to-state mapping. We show that this constraint restricts temporal capacity independently of gradient stability. To address this, we introduce structured high-dimensional state dynamics that preserve temporal separability and prevent saturation of the input-to-state Jacobian over time. On controlled sequence tasks, the resulting model exhibits rapid learning despite remaining strictly within a standard RNN formulation. Our results demonstrate that RNN performance is governed by the evolution and preservation of temporal rank, and suggest that mitigating gradient pathologies alone is insufficient without explicitly maintaining temporal capacity.
\end{abstract}

\section{Introduction}
Recurrent neural networks (RNNs) are non-linear dynamical systems for modelling sequential data, in which a temporally evolving hidden state is updated via discrete-time inputs and parameters. In their simplest forms, such as Elman \cite{elman90} and Jordan \cite{jordan86} networks, RNNs are Turing-complete under certain assumptions \cite{siegelmann95, chung21}; that is, they can compute any function that a general-purpose computer can.

Despite their theoretical expressivity, simple RNNs trained via back-propagation through time (e.g. \cite{robinson87}) are known to suffer from \textit{vanishing} and \textit{exploding} gradients \cite{hochreiter91,bengio94,hochreiter01}, a key motivation behind the development of more complex designs such as the Long Short-Term Memory (LSTM) \cite{hochreiter97,gers99} and Gated Recurrent Unit (GRU) \cite{cho2014}.

Although gradient instability can be mitigated through training-time countermeasures such as gradient clipping \cite{pascanu13} and careful weight initialisation \cite{glorot10,saxe14,mikolov15,he15,le15}, as well as structural approaches including gating \cite{hochreiter97,cho2014} and parametrisation \cite{arjovsky15,mhammedi17,helfrich18}, recurrent architectures have nevertheless been supplanted by architectures better suited to parallel computation. Transformers \cite{vaswani17} have filled this gap and advanced the state-of-the-art in sequence modelling \cite{radford18,gulati20,dosovitskiy21}: by eliminating recurrence, they allow modern hardware to process entire sequences simultaneously, improving training efficiency and scalability despite prior efforts to accelerate recurrent models \cite{he17,lei18}.

A major drawback of Transformers is their quadratic time and memory complexity, as self-attention computes pairwise interactions between all inputs. This has motivated two broad lines of work: improving efficiency through specialised hardware-aware algorithms \cite{dao22,dao23,shah24,zadouri26} and reducing complexity via architectural modifications \cite{tay22a,parmar18,child19,kitaev20,roy20,dai20,tay22b,wang20,katharopoulos20}.

Notably, a kernel-based approach by \citeauthor{katharopoulos20} \cite{katharopoulos20} admits a linear recurrent formulation. Similar models, including linear RNNs (LRNNs) \cite{orvieto23} and state space models such as S4 and Mamba \cite{gu20,gu22,gu24}, scale linearly with sequence length and can be parallelised efficiently \cite{martin18}. However, prior work suggests that, from a circuit complexity viewpoint, both LRNNs and Transformers are less expressive than general RNNs, particularly on tasks involving unbounded state tracking \cite{deletang23,merrill25,merrill26}.

Given this shortcoming, there has been a resurgence of interest in RNNs, driven both by hardware-efficient algorithms \cite{poppel25} and parallelisation strategies \cite{wang21,lim24,gonzalez25,danieli25}, as well as new architectural variants such as xLSTM \cite{beck24}.

While these developments address optimisation and efficiency, recurrent models can also be reconsidered through the lens of their hidden state evolution. Beyond vanishing and exploding gradients, which have largely dominated prior approaches to understanding and mitigating long-range dependency issues, we adopt a complementary viewpoint based on the rank structure induced by this evolution, characterising information propagation in terms of the number of independent input directions that can be preserved over time.

Building on this perspective, we propose a deliberately structured, non-parametric formulation of recurrent dynamics that sustains the growth of stable rank over time, and use it to demonstrate that preserving this capacity for continued rank growth is important for efficient learning and long-range information propagation.

The rest of the paper is organised as follows. We first formalise a rank-based perspective on recurrent hidden state evolution and its implications for information propagation and learning dynamics. We then analyse how state dimensionality and initialisation influence this temporal capacity. Finally, we evaluate these ideas on controlled sequence modelling tasks.

\section{Background}

A Recurrent Neural Network (RNN) is a discrete-time non-linear dynamical system of the form $\mathbf{h}_t = \Phi(\mathbf{h}_{t-1},\mathbf{x}_t)$ that takes inputs $\mathbf{x}_1, \mathbf{x}_2,... \in \mathbb{R}^n$ sequentially and defines a trajectory map
\begin{equation}
    F^{\theta}(\mathbf{x}_{1:T}) \coloneqq \mathbf{h}_T
\end{equation}
for some total sequence length $T$ and constant embedding size $d$. For fixed parameters $\theta = (\mathbf{U}, \mathbf{W}, \mathbf{b})$ and initial state $\mathbf{h}_0 \in \mathbb{R}^d$, a typical Elman RNN \cite{elman90} defines $\Phi$ via
\begin{equation}
    \label{eq:elman}
    \Phi(\mathbf{h},\mathbf{x}) = \phi(\mathbf{U}\mathbf{h} + \mathbf{W}\mathbf{x} + \mathbf{b})
\end{equation}
where $\phi(\cdot)$ is an element-wise non-linear activation function, $\mathbf{U} \in \mathbb{R}^{d \times d}$, $\mathbf{W} \in \mathbb{R}^{d \times n}$, and $\mathbf{b} \in \mathbb{R}^d$. An output for each time step is given by
\begin{equation}
    \mathbf{y}_t = \phi_y(\mathbf{V}\mathbf{h}_t + \mathbf{b}_y)
\end{equation}
with $\mathbf{V} \in \mathbb{R}^{m \times d}, \mathbf{b}_y \in \mathbb{R}^m$.

To characterise how input sequences influence the hidden state, we consider the Jacobian of $\mathbf{h}_t$ with respect to the input sequence, defined as the concatenation
\begin{equation}
    \label{eq:jac}
    \mathbf{J}_t = \left[\frac{\partial \mathbf{h}_t}{\partial \mathbf{x}_1} \frac{\partial \mathbf{h}_t}{\partial \mathbf{x}_2} \cdots \frac{\partial \mathbf{h}_t}{\partial \mathbf{x}_t}\right] \in \mathbb{R}^{d \times nt}
\end{equation}
Each block $\frac{\partial \mathbf{h}_t}{\partial \mathbf{x}_\tau}$ captures the sensitivity of the hidden state at time $t$ to the input at time $\tau$, and their concatenation aggregates these dependencies across time, enabling analysis of the effective dimensionality of input directions influencing the hidden state.

We analyse the dimensionality of $\mathbf{J}_t$ using both rank and stable rank \cite{rudelson06}. For a matrix $\mathbf{A}$, the stable rank is defined as
\begin{equation}
    \operatorname{srank}(\mathbf{A})= \frac{\|\mathbf{A}\|_F^2}{\|\mathbf{A}\|_2^2} = \frac{\sum_i \sigma_i^2}{\sigma^2_1}
\end{equation}
where $\|\cdot\|_F$ and $\|\cdot\|_2$ denote the Frobenius and spectral norms, respectively, and $\{\sigma_i\}$ are the singular values of $\mathbf{A}$ with $\sigma_1 \ge \sigma_2 \ge \cdots$. Stable rank provides a continuous proxy for rank that captures effective dimensionality and is less sensitive to small singular values. It holds that $\operatorname{srank}(\mathbf{A}) \le \operatorname{rank}(\mathbf{A})$, with equality when all non-zero singular values are equal.

We now use these tools to analyse the rank and stable rank of $\mathbf{J}_t$ over time in RNNs.

\section{Rank Analysis of RNNs}

\begin{prop}
    \label{prop:cap}
    For an Elman RNN defined by \eqref{eq:elman}, the Jacobian of the hidden state $\mathbf{h}_t$ with respect to the input sequence $(\mathbf{x}_1,...,\mathbf{x}_t)$ satisfies
    \begin{equation}
        \operatorname{rank}(\mathbf{J}_t) \le d \quad \forall t \le T
    \end{equation}
\end{prop}

\begin{proof}
    The rank of a matrix is upper-bounded by its number of rows. Since $\mathbf{J}_t \in \mathbb{R}^{d \times nt}$, it follows that $\operatorname{rank}(\mathbf{J}_t) \le d$.
\end{proof}

This trivial bound highlights that the effective input dimensionality is constrained by the hidden state dimension and holds independently of the choice of non-linearity $\phi$, reflecting a structural limitation imposed by the fixed hidden state size.

In this context, stable rank can be interpreted as the effective number of dominant singular directions, and thus as the effective dimensionality of input perturbations that meaningfully influence the state at time $t$.

\begin{remark}
    Although bounded above by $d$ (Proposition~\ref{prop:cap}), we observe that the stable rank of $\mathbf{J}_{t}$ is often much smaller than $d$ in practice.
\end{remark}

\begin{figure}
    \centering
    \includegraphics[width=\linewidth]{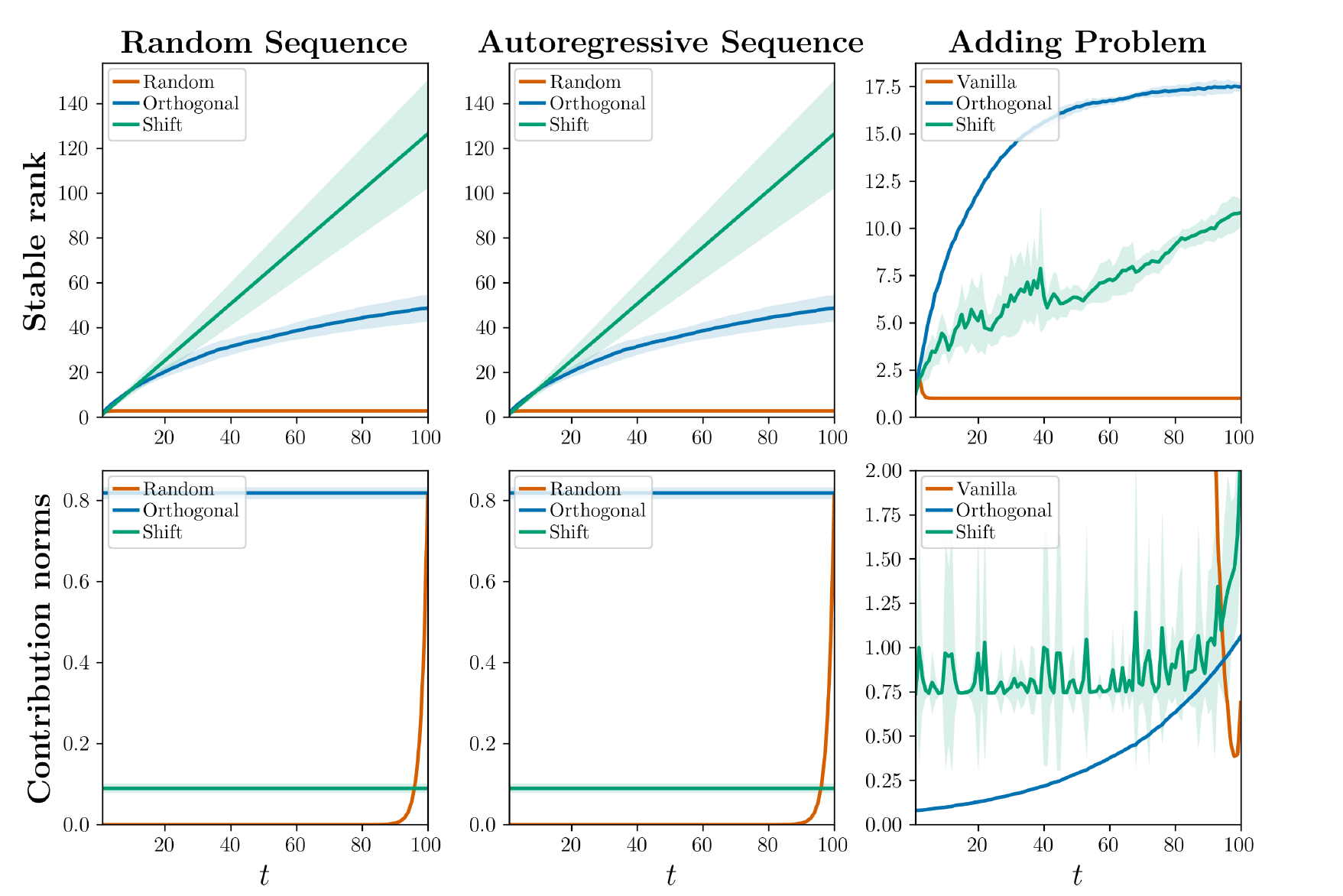}
    \caption{Stable rank and input contribution norms to the final hidden state $\mathbf{h}_T$ across random, autoregressive, and adding sequence tasks for different recurrent parametrisations.}
    \label{fig:rank}
\end{figure}

Details on specific cases are provided in Appendix~\ref{app:rank}. Figure~\ref{fig:rank} demonstrates this behaviour under different recurrent parametrisations (including the proposed shift structure, introduced in Section~\ref{sec:prop}).

We plot the contribution norms $\|\frac{\partial \mathbf{h}_T}{\partial \mathbf{x}_t}\|$, quantifying the sensitivity of the final hidden state to perturbations at each time step.

For a comprehensive view, we consider random inputs, a structured autoregressive sequence, and trained models on the adding problem. All models have the same hidden dimension, $d=200$, and are run for $100$ time steps.

First, we examine the dynamics of untrained RNNs under different input regimes. Across both random sequences and structured autoregressive inputs, the stable rank of $\mathbf{J}_t$ remains significantly below its ambient dimension, and saturates rapidly over time, with a delayed onset in the case of the orthogonally-constrained recurrent matrix. In the vanilla RNN, this behaviour is accompanied by concentrated contribution norms, suggesting that only a small number of recent time steps dominate the sensitivity of the final hidden state.

We then consider trained models on the adding task. With the introduction of non-linearity, the vanilla RNN exhibits stable contribution norms for recent inputs (up to approximately ten time steps in the past), followed by a sharp increase for earlier inputs, indicating highly unstable long-range sensitivity. While the orthogonally-constrained RNN maintains more controlled contribution magnitudes over time, its stable rank still plateaus well below the ambient dimension, and neither model is able to solve the task.

\paragraph{Vanishing \& Exploding Gradients} Writing the input-to-state Jacobian as $\mathbf{J}_t$, as defined in \eqref{eq:jac}, classical analyses of vanishing and exploding gradients focus on the magnitudes of back-propagated signals, which in this formulation correspond to the norms of the columns of $\mathbf{J}_t$. In contrast, we study the geometry of these sensitivity vectors.

Even when gradient norms are preserved (e.g., under orthogonal recurrent dynamics), the corresponding sensitivity vectors may become increasingly aligned over time, leading to a low (stable) rank of $\mathbf{J}_t$. This aligns with our empirical observation that stable rank can plateau even when contribution norms remain well-conditioned, indicating that well-conditioned gradient norms do not preclude collapse in the effective dimensionality.

A low stable rank implies that sensitivities to past inputs concentrate in a low-dimensional subspace. For tasks requiring the preservation of multiple independent input directions, this presents an inherent limitation under a fixed hidden dimension. Even when the task is effectively low-dimensional, challenges remain: contributions from different time steps tend to align within this subspace, reducing their separability and making credit assignment during back-propagation ill-conditioned. This phenomenon is distinct from, but complementary to, vanishing and exploding gradients. This motivates architectures that explicitly control or preserve directional diversity over time.

\section{Proposed Construction}
\label{sec:prop}

To reduce alignment of input directions, we consider a structured expansion of the recurrent state in which temporal information is explicitly distributed across a higher-dimensional state space indexed by the sequence horizon $T$. To this end, we augment the hidden state to $\mathbf{h} \in \mathbb{R}^{Td'}$ for some per-step embedding size $d'$ such that the total hidden dimension is $d = Td'$ (necessitating $\mathbf{U} \in \mathbb{R}^{Td' \times Td'}$, $\mathbf{W} \in \mathbb{R}^{Td' \times n}$, and $\mathbf{b} \in \mathbb{R}^{Td'}$).

$\mathbf{U}$ is non-parametric, shifting each $d'$-dimensional block down by one position via an off-diagonal identity, while $\mathbf{W}$ and $\mathbf{b}$ are zero-padded so that only the top $d'$-dimensional block receives new input, meaning lower blocks evolve purely through the recurrent shift and non-linearity. Formalising this:
\begin{equation} 
    \label{eq:mats} 
    \mathbf{U}_{ij} = \begin{cases} 
        1 & \text{if } i = j + d' \\
        0 & \text{otherwise} 
    \end{cases}
    \quad
    \mathbf{W}_{ij} = \begin{cases} 
        w_{ij} & \text{if } 1 \le i \le d' \\
        0 & \text{otherwise}
    \end{cases}
    \quad
    \mathbf{b}_i = \begin{cases}
    b_i & \text{if } 1 \le i \le d' \\
    0 & \text{otherwise}
    \end{cases}
\end{equation}
Despite this structured expansion, the model remains an Elman RNN, with a fixed, non-parametric recurrent operator. The structure naturally raises the question of whether repeated application of non-linearity preserves information and avoids destructive alignment in the expanded hidden state, which is explored, along with some inversion theory, in Appendix~\ref{app:inv}.

\begin{remark} 
    The full $\mathbf{U}$ matrix in \eqref{eq:mats} is conceptually of size $Td' \times Td'$. However, the hidden state can be implemented via an explicit shift of $d'$-dimensional blocks, avoiding the need to materialise either the full $Td'$-dimensional state or $\mathbf{U}$, and allowing $T$ to remain unspecified at construction time.

    Analogous sparse implementations are used for the input and bias parameters. A detailed description of the resulting computational and memory complexity is provided in Appendix~\ref{app:imp}.
\end{remark}

For random and autoregressive inputs, the shift-structured model exhibits uniform contribution norms, similar in appearance to those induced in the orthogonally-constrained model. However, in contrast to both standard and orthogonal RNNs, the stable rank of $\mathbf{J}_t$ increases linearly over time, with no evidence of saturation, and remains substantially higher than in the orthogonally-constrained model (Figure~\ref{fig:rank}).

On the adding problem, the shift RNN initially exhibits a more uniform and temporally distributed stable rank allocation; however, after training, the dynamics reorganise to reflect the task structure, leading to a distinct evolution of stable rank over time.

\begin{figure}
    \centering
    \includegraphics[width=\linewidth]{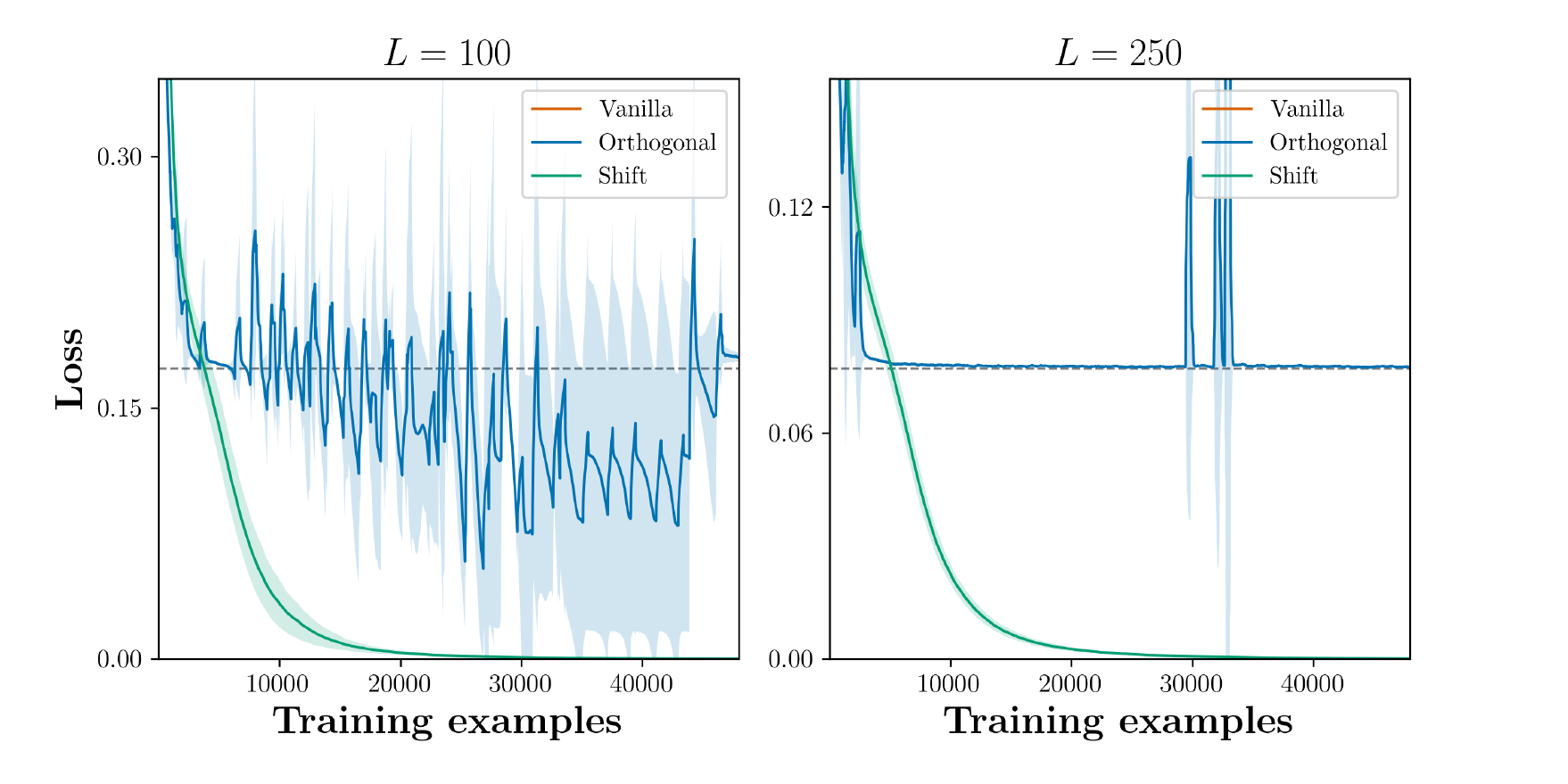}
    \caption{Performance on the copy task for lags $L = 100$ and $L = 250$: the vanilla RNN diverges and falls outside the plotting range (hence not visible in the figure); the orthogonal RNN briefly approaches near-perfect performance at $L = 100$ but ultimately returns to the baseline and does not outperform it at $L = 250$, while the shift RNN solves both settings.}
    \label{fig:copy}
\end{figure}

\section{Experiments}
\label{sec:exp}

We now evaluate the construction on two canonical sequence reasoning tasks — the copy task and the adding problem — which are widely used to probe long-range memory in recurrent neural networks \cite{hochreiter97,arjovsky15,wisdom16,helfrich18}. While these tasks are often discussed in terms of vanishing and exploding gradients, we show that this perspective is incomplete: the evolution of the stable rank of the input-to-state mapping over time provides a complementary and informative characterisation of model behaviour.

To test this, we compare three single-layer recurrent architectures: a standard RNN, an orthogonal RNN, and our shift construction. All models are assigned the same hidden state dimension $d$, which is scaled linearly with sequence length as $d = Td'$. In our experiments, we fix $d' = 8$ (yielding $d=8T$), and we use the ELU activation \cite{clevert16} (for more results with other non-linearities, see Appendix~\ref{app:res}). This controlled setting allows us to isolate the effects of the recurrent dynamics on representational structure, independent of depth or architectural gating mechanisms. All results are averaged over $3$ runs with different random seeds, with standard deviation reported in the plots.

\paragraph{Copy Task} The copy task was introduced in \cite{hochreiter97} and we follow the formulation of \cite{arjovsky15}. Each sequence consists of $10$ data tokens (e.g. digits $0$-$7$) at the beginning, followed by a time lag ($L$) of blank tokens and a delimiter, after which the model is required to output the original $10$-token sequence. The model must output the blank token for all other time steps. The total sequence length is $L + 20$. The baseline performance is given by outputting the blank tokens followed by uniform guesses over the 8-symbol alphabet for 10 steps, yielding a cross-entropy loss of $\frac{10 \ln 8}{L+20}$. See Figure~\ref{fig:copy} for results.

For time lags of $L = 100$ and $L = 250$, the shift RNN is the only one that consistently outperforms the baseline and successfully solves the task. The orthogonal RNN briefly approaches near-perfect performance with $L = 100$, but fails to maintain this behaviour and does not outperform the baseline at $L = 250$. In contrast, the standard RNN diverges in these settings. Even when the hidden state dimension is scaled with sequence length, both baseline models fail to solve the task across the evaluated regimes.

\begin{figure}
    \centering
    \includegraphics[width=\linewidth]{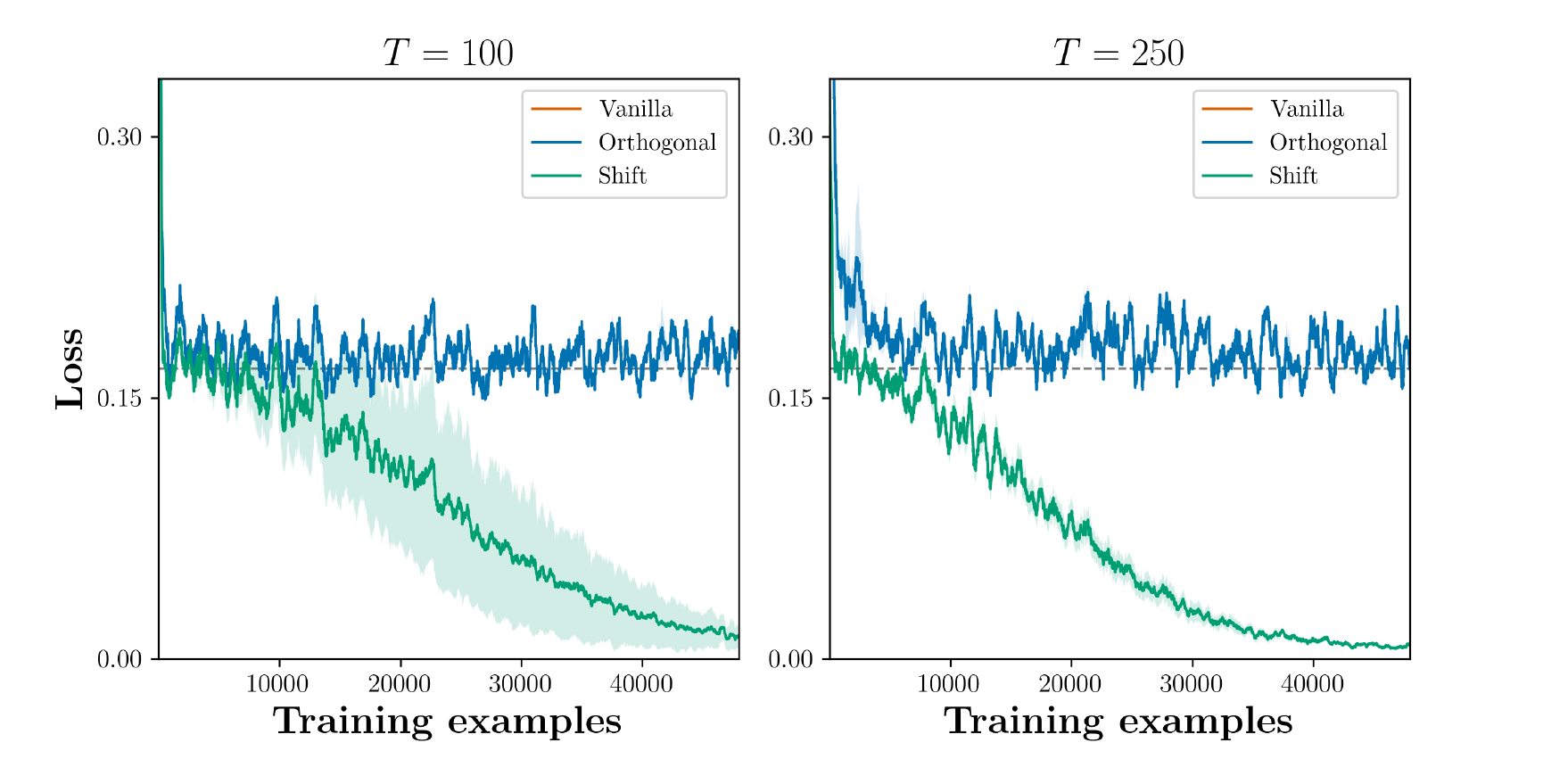}
    \caption{Performance on the adding problem for sequence lengths $T = 100$ and $T = 250$: the vanilla RNN diverges and falls outside the plotting range (hence not visible in the figure); the orthogonal RNN does not outperform the baseline, while the shift RNN solves the task.}
    \label{fig:add}
\end{figure}

\paragraph{Adding Problem} The adding problem is another task defined in both \cite{hochreiter97} and \cite{arjovsky15}, and we opt for the latter's formulation. The input consists of a sequence of pairs: the first element is a real value sampled from $\mathcal{U}[0,1]$, and the second is a binary marker in $\{0,1\}$. The aim is to output the sum of the two real values where the marker equals $1$; these positions are sampled uniformly at random, one located in the first half of the sequence and one in the latter half. The baseline performance is given by a naïve output of $1$ regardless of the sequence, yielding a mean-squared-error loss of $0.167$. See Figure~\ref{fig:add} for results.

Once again, the vanilla RNN is unstable and diverges in these settings. The orthogonally-constrained RNN does not outperform the baseline on this task, while the shift RNN consistently achieves successful performance at both sequence lengths.

We provide additional results for the copy task and adding problem using the shift RNN in Appendix~\ref{app:res}. We also note that the standard RNN does not always diverge under all activation functions; these extended results are included in the appendix.

\section{Discussion}

We have analysed rank evolution as a structural constraint in RNNs, showing that the inputs-to-state Jacobian in both vanilla and orthogonal RNNs exhibits stable rank saturation well below the ambient dimension. This provides a complementary perspective to gradient-based analyses: even in regimes where gradients are well-behaved, recurrent dynamics can collapse input directions, limiting the effective dimensionality of representations over time. In this sense, vanishing and exploding gradients are not sufficient to characterise the challenges of learning long-range dependencies.

Our construction demonstrates that this limitation is not inherent to recurrence per se, but to how representational capacity is allocated across time. By enforcing a shift structure, the model avoids rank saturation and instead distributes representational capacity over the sequence. While this construction is deliberately simple, it serves as a proof-of-principle that stable rank can be preserved and accumulated over time within a standard RNN formulation.

Taken together, these results suggest a design principle: sequence models should explicitly manage how representational rank is allocated across inputs. Architectures such as Transformers and convolutional sequence models implicitly follow this principle through attention and receptive fields, which provide structured mechanisms for preserving input distinctions. In contrast, standard recurrent architectures rely on mixing dynamics, which can lead to premature saturation of stable rank. Understanding how common mechanisms such as gating affect rank allocation over time remains an important direction for future work.

Although the proposed architecture is memory-efficient because the full recurrent matrix need not be instantiated, ``mixing'' of elements in the sequence is not explicitly handled. Whether this is a core and persistent tension when using RNNs needs to be investigated, and we leave this to future work: one may introduce controlled mixing mechanisms that preserve stable rank over time.

Finally, the dependence on sequence length raises computational concerns, but also suggests a potential advantage: when task-relevant input directions are not known a priori, an over-complete, time-expanded representation can facilitate learning by deferring compression. This motivates a two-stage perspective in which models first construct a high-rank representation distributed over time and subsequently compress it into a task-relevant subspace, either during training or post-hoc. Developing architectures that approximate this behaviour is a promising direction for future work.

\bibliography{refs}

\appendix

\section{Probing the Saturation of Stable Rank Under Assumptions}
\label{app:rank}

\subsection{Experiment Details}

We use \texttt{torch} for our implementations, and consider three input regimes to test the robustness of the temporal rank results:
\paragraph{Random} i.i.d. random inputs are sampled as independent Gaussian noise:
\begin{equation}
    \mathbf{x}_t \sim \mathcal{N}(0,I)
\end{equation}

\paragraph{Autoregressive} Temporally correlated sequences are generated via a first-order autoregressive (AR(1)) process:
\begin{equation}
    \mathbf{x}_t = \rho\mathbf{x}_{t-1} + \sqrt{1-\rho^2}\epsilon_t, \quad \epsilon_t \sim \mathcal{N}(0,I)
\end{equation}
with $\rho = 0.9$ to ensure strong temporal dependence while maintaining stationarity. The initial state is sampled as $\mathbf{x}_0 \sim \mathcal{N}(0,I)$.

\paragraph{Adding Problem} Follows the definition provided in Section~\ref{sec:exp}, where each model is trained with $48{,}000$ examples. Inputs consist of scalar values from $\mathcal{U}[0,1]$ with binary markers indicating two positions (one uniformly at random in the first half, one uniformly at random in the latter half), where the target is their sum. Models use the ELU activation function.

The following results in this section use random input sequences with randomly initialised models, as earlier results have shown that pre-training behaviour is not sensitive to input structure. We also use a linear activation unless stated otherwise.

\subsection{Singular Value Spectra}

\begin{figure}
    \centering
    \includegraphics[width=\linewidth]{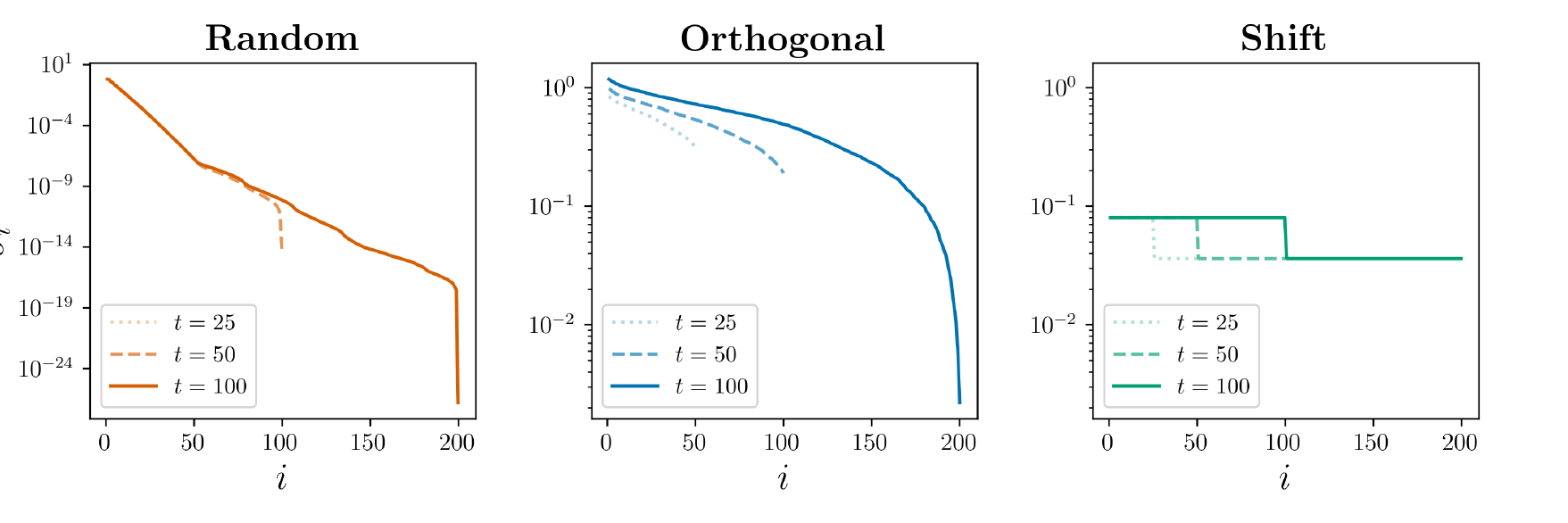}
    \caption{Singular value spectra of the input-to-state Jacobian at different time steps: vanilla RNN exhibits rapid collapse, orthogonal RNN shows gradual skewing over time, while the shift RNN maintains a near-flat spectrum with a structured step.}
    \label{fig:spec}
\end{figure}

We provide additional plots of the singular value spectra for different time steps across models in Figure~\ref{fig:spec}, complementing stable rank analysis by revealing how these values are distributed.

The randomly initialised RNN clearly suffers from spectral collapse into a few dominant modes. Plotted on logarithmic scale, the spectrum exhibits an approximately exponential decay, with most singular values rapidly approaching numerical zero. The spectra at the different time steps largely overlap, indicating collapse occurs quickly and changes little thereafter.

The orthogonally initialised RNN preserves the overall scale of singular values, but the spectrum still becomes progressively skewed over time, with increasing decay in the tail at later time steps.

The shift RNN exhibits a near-flat spectrum with a clear step structure, showing two levels of uniformly preserved singular values. The location of the step scales with time steps, reflecting explicit allocation of the independent directions over time.

While this analysis provides a geometric view of the Jacobian structure, our main focus remains on stable rank as a compact measure of effective dimensionality.

\subsection{Sequence Length Variation}

\begin{figure}
    \centering
    \includegraphics[width=\linewidth]{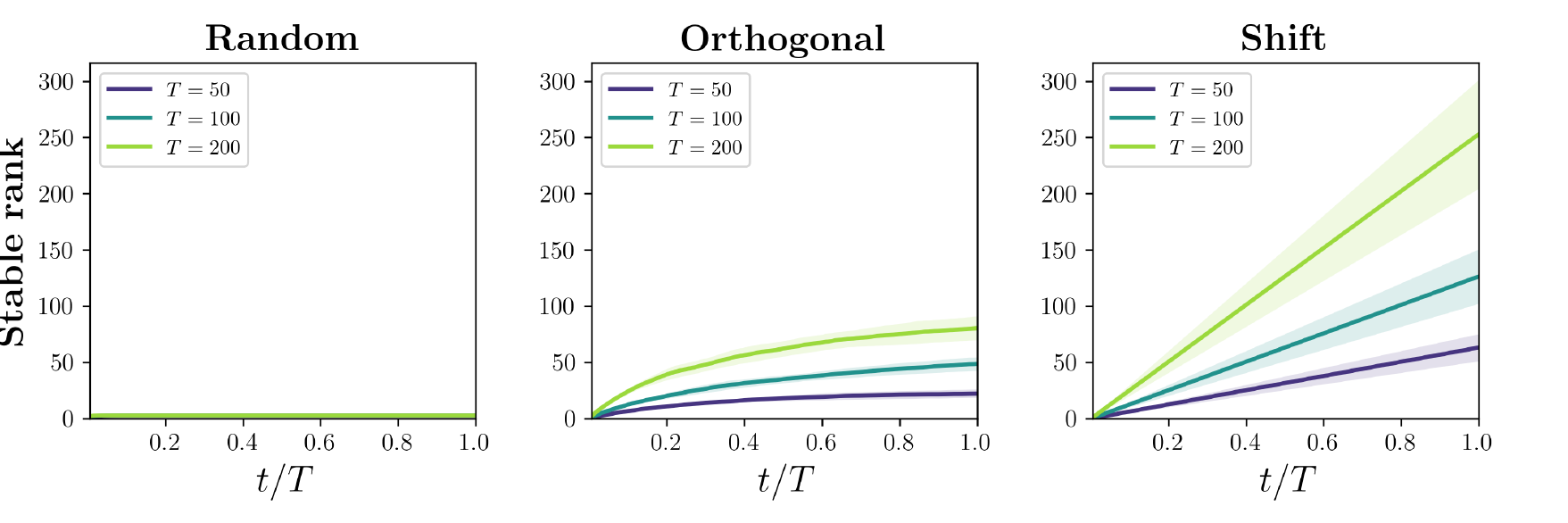}
    \caption{Stable rank evolution over normalised time $\frac{t}{T}$ for varying sequence lengths; only the shift RNN exhibits near-linear scaling of stable rank with $T$, while orthogonal shows diminishing gains and vanilla remains collapsed.}
    \label{fig:seqlen}
\end{figure}

In Figure~\ref{fig:seqlen}, we evaluate stable rank as a function of normalised time $\frac{t}{T}$ across varying sequence lengths $T \in \{50,100,200\}$, with hidden dimension scaled proportionally to maintain comparable capacity across settings. We observe that only the shift RNN exhibits near-linear scaling of stable rank with $T$, approximately doubling when $T$ is doubled. In contrast, the randomly initialised RNN remains flat across all settings, while the orthogonal RNN shows improved scaling but with diminishing returns as $T$ increases.

\subsection{Hidden Dimension Variation}

\begin{figure}
    \centering
    \includegraphics[width=\linewidth]{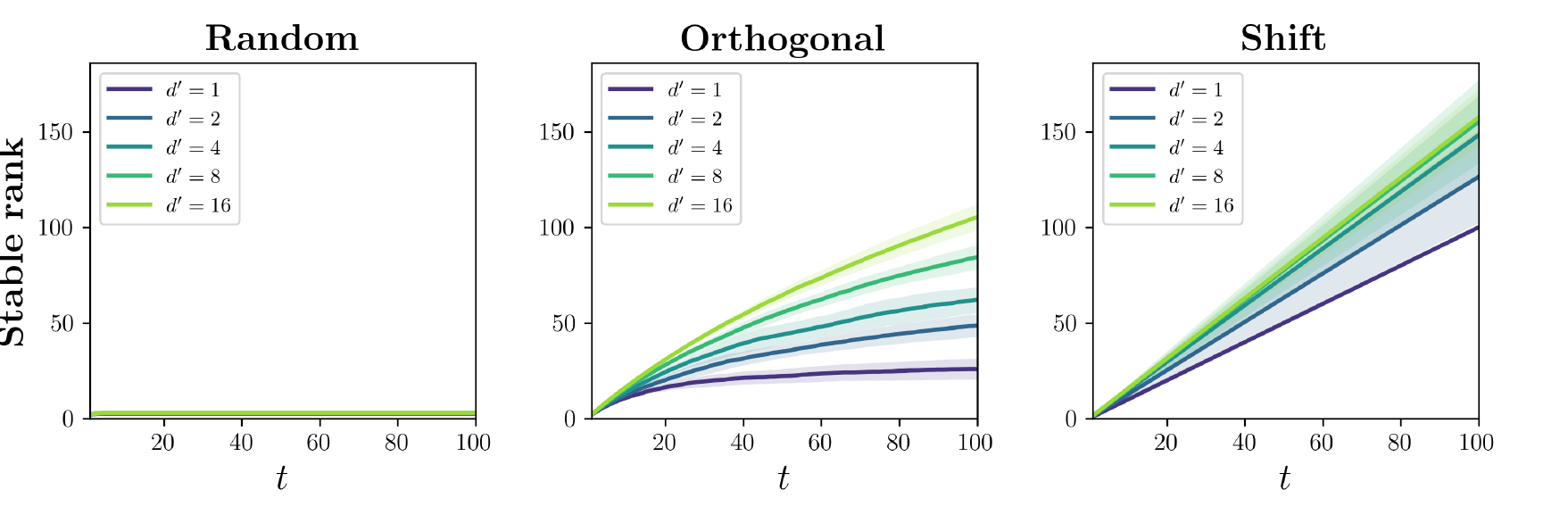}
    \caption{Effect of hidden dimension on stable rank dynamics; the shift RNN achieves efficient allocation at low dimensions, while the orthogonal RNN requires substantially larger hidden states to match performance.}
    \label{fig:dim}
\end{figure}

In Figure~\ref{fig:dim}, we further study the effect of hidden dimension on stable rank dynamics, setting $d = Td'$ for $d' \in \{1,2,4,8,16\}$. The randomly initialised RNN remains flat across all dimensions, indicating no effective temporal allocation of representational capacity. The orthogonal RNN benefits strongly from increased hidden dimension, with $d' = 16$ achieving stable rank levels comparable to the shift RNN with $d' = 1$. However, its gains diminish with increasing $T$, suggesting that substantially larger hidden states are required to maintain effective stable rank allocation over longer time horizons. In contrast, the shift RNN exhibits stronger efficiency at low dimensions, but shows diminishing returns as $d'$ increases, perhaps consistent with the time-to-dimension allocation structure.

\subsection{Non-Linearity Variation}

\begin{figure}
    \centering
    \includegraphics[width=\linewidth]{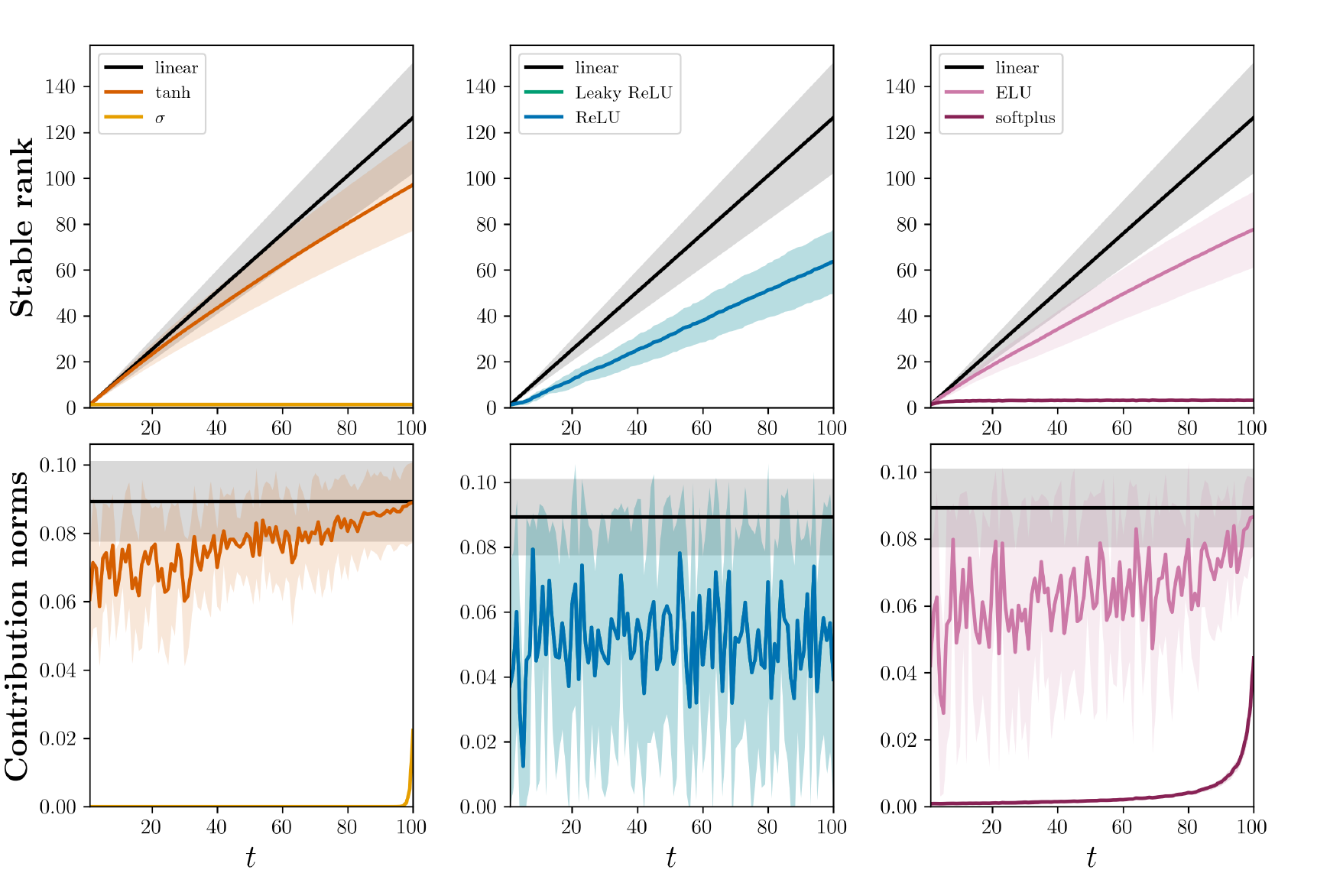}
    \caption{Grouped non-linear activation functions and their effect on stable rank and contribution norms in the shift RNN with a linear control; $\tanh$ yields the highest stable rank, followed by ELU and ReLU variants, while sigmoid and softplus lead to pronounced collapse.}
    \label{fig:nl}
\end{figure}

In Figure~\ref{fig:nl}, we evaluate the effect of non-linearities using the shift RNN, with the linear case serving as a control baseline. We group non-linearities into three categories:
\begin{enumerate*}[label=(\roman*)]
    \item saturating ($\tanh$, logistic sigmoid),
    \item piecewise-linear (ReLU, leaky ReLU), and
    \item smooth, non-saturating (ELU, softplus),
\end{enumerate*}
to compare their effects on temporal rank dynamics.

ReLU and leaky ReLU exhibit nearly identical behaviour in both stable rank and contribution norm analyses, so we treat them as equivalent in this setting. Among non-linearities, $\tanh$ yields the highest stable rank and the most uniform distribution of contribution norms, although its growth rate diminishes over time. ELU performs almost as well, while both ReLU variants underperform this in comparison. As expected, sigmoid exhibits strong collapse due to saturation effects; interestingly, softplus also shows a similar collapse, suggesting that smoothness alone is insufficient to preserve temporal rank.

\subsection{Variation Summary}

Across all variations, the shift RNN consistently provides the most effective allocation of temporal rank over time, but achieves this through a highly structured, non-mixing dynamics that may be restrictive in practice. This suggests the need to identify broader classes of recurrent transformations that can preserve temporal rank allocation while retaining expressive mixing. Non-linearities can be incorporated into this framework, but generally reduce stable rank relative to the linear shift case. In contrast, in the fully unconstrained vanilla RNN setting — where mixing is maximised — the differences between non-linearities are largely negligible due to rapid and uniform collapse of the Jacobian structure.

\section{Efficient Realisation and Computational Complexity}
\label{app:imp}

The proposed recurrent construction is defined in terms of a conceptual state-transition operator $\mathbf{U} \in \mathbb{R}^{Td' \times Td'}$ and a hidden state $\mathbf{h}_t \in \mathbb{R}^{Td'}$. However, neither $\mathbf{U}$ nor the full dense parametrisation is explicitly instantiated. Instead, $\mathbf{U}$ is realised as a deterministic shift operator acting on $d'$-dimensional blocks of the hidden state.

Each update consists of a block-wise shift, input injection into the top $d'$ coordinates, and element-wise application of the nonlinearity over the full $Td'$-dimensional state. The input and bias parameters are conceptually given by $\mathbf{W} \in \mathbb{R}^{Td' \times n}$ and $\mathbf{b} \in \mathbb{R}^{Td'}$, but only their first $d'$ coordinates are non-zero and explicitly stored. The output layer is a standard linear readout $\mathbf{V} \in \mathbb{R}^{m \times Td'}$ with bias $\mathbf{b}_y \in \mathbb{R}^{m}$.

\paragraph{Per-step complexity.}
Let $T$ denote the sequence length and $d'$ the block size, so that the hidden dimension is $Td'$. The dominant per-step costs are block shift, nonlinearity, and readout, giving
\begin{equation}
    \mathcal{O}(Td' + nd' + mTd')
\end{equation}

\paragraph{Sequence complexity.}
Over a sequence of length $T$, the total cost depends on the output regime:

\begin{itemize}
    \item \textbf{Step-wise output (e.g. copy task):}
    outputs are computed at every time step, giving
    \begin{equation}
        \mathcal{O}(T^2 d'(1+m) + T n d').
    \end{equation}

    \item \textbf{Final-step output (e.g. adding problem):}
    only the final hidden state is read out, giving
    \begin{equation}
        \mathcal{O}(T^2 d' + mTd' + T n d').
    \end{equation}
\end{itemize}

\paragraph{Memory complexity.}
The model stores a hidden state of size $\mathcal{O}(Td')$, sparse input and bias parameters of size $\mathcal{O}(nd')$ and $\mathcal{O}(d')$, and a readout matrix of size $\mathcal{O}(mTd')$.

\section{Recovering Inputs}
\label{app:inv}

Given that the shift RNN allocates distinct subspaces across time, a natural question is whether this structure permits the recovery of past inputs from the hidden state. We explore this briefly by analysing the invertibility of non-linear activations, as well as practical methods of recovery.

\subsection{On the Effect of Repeated Non-Linearity Application}

\begin{wrapfigure}[21]{r}{0.5\textwidth}
    \centering
    \includegraphics[width=\linewidth]{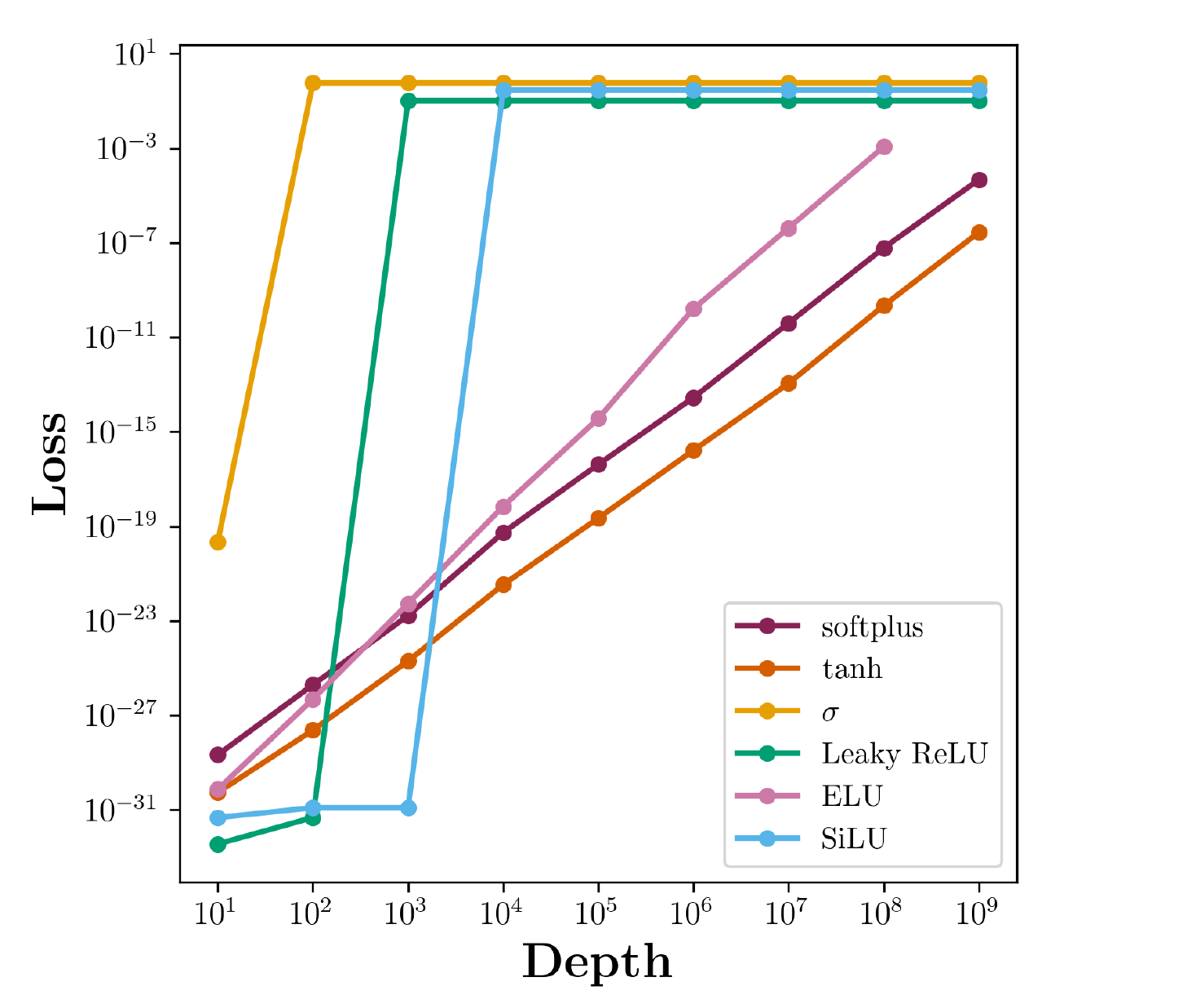}
    \caption{Inverting non-linearities on continuous values taken from the distribution $\mathcal{U}[-1,1]$ up to depths of one billion.}
    \label{fig:inv}
\end{wrapfigure}

Not all non-linearities commonly used in RNNs, e.g. ReLU \cite{le15}, are invertible. Of those that are, repeated application at large depths may lead to numerical precision issues. In particular, bounded non-linearities such as the logistic sigmoid saturate after relatively few iterations, leading to loss of information under finite precision arithmetic. While $\tanh$ is also bounded and formally invertible, values near the boundaries at $-1$ and $1$ become increasingly unstable, and small numerical errors can lead to large deviations under repeated inversion.

In the experiment in Figure \ref{fig:inv}, we apply common non-linearities on $100$ i.i.d. real values drawn from $\mathcal{U}[-1,1]$ up to depths of one billion. The non-linearities were implemented using Python's native \texttt{math} library operating on scalar \texttt{float64} values without relying on tensor libraries or automatic numerical stabilisation beyond that provided by \texttt{math}. No clamping or special-case handling was introduced.

Although Leaky ReLU is piecewise linear, repeated multiplication by the slope factor leads to degradation in recovery beyond moderate depths (around $10^2$), consistent with finite-precision underflow effects. The logistic sigmoid rapidly saturates, with information loss occurring after approximately $10$ iterations. SiLU, defined as $x \cdot \sigma(x)$, delays this collapse to larger depths (around $10^3$). In contrast, $\tanh$ remains numerically stable over significantly larger depths, with error increasing more gradually. Non-linearities such as softplus and ELU also remain stable across most depths; however, at extreme depths (around $10^9$), ELU exhibits a late-stage numerical breakdown.

\subsection{Full Recovery Theory}

\begin{prop}
    Following the construction provided in Section~\ref{sec:prop}, assume:
    \begin{enumerate}
        \item $\phi$ is invertible
        \item $n=d'$ and $\mathbf{W}_{d'} \in \mathbb{R}^{d' \times d'}$ is invertible.
    \end{enumerate}
    Then at the final time step $T$, any input $\mathbf{x}_1,...,\mathbf{x}_T$ can be exactly recovered from the hidden state $\mathbf{h}_T$.
\end{prop}

\begin{proof}
    To decode the input sequence from the final state, first partition the final state into $T$ slices of length $d'$, $\mathbf{h}^{(T)}_T,...,\mathbf{h}^{(1)}_T$. Due to the shift structure, $\mathbf{h}^{(t)}_T$ is simply the hidden state calculated from the input embedding at time $t$, $\mathbf{h}^{(t)}_{t}$, with non-linearity $\phi$ applied $T-t$ times. Denoting the parametric portions of $\mathbf{W}$ and $\mathbf{b}$ as $\mathbf{W}_{d'} \in \mathbb{R}^{d' \times d'}$ and $\mathbf{b}_{d'} \in \mathbb{R}^{d'}$, we take $\mathbf{h}^{(T)}_T$ and solve:
    \begin{equation}
        \mathbf{h}^{(T)}_T = \phi(\mathbf{W}_{d'}\mathbf{x}_T + \mathbf{b}_{d'})
    \end{equation}
    for $\mathbf{x}_T$. If $\phi$ is invertible, $\mathbf{W}_{d'}\mathbf{x}_T$ can be computed exactly via:
    \begin{align}
        \mathbf{W}_{d'}\mathbf{x}_T 
        &= \phi^{-1}(\mathbf{h}^{(T)}_T)
        - \mathbf{b}_{d'} \\
        \Rightarrow \quad
        \mathbf{x}_T 
        &= \mathbf{W}_{d'}^{-1} \Big[
        \phi^{-1}(\mathbf{h}^{(T)}_T)
        - \mathbf{b}_{d'}
        \Big]
    \end{align}
    We can generalise the above to any input $\mathbf{x}_t$ by:
    \begin{align}
        \label{eq:rec}
        \mathbf{W}_{d'}\mathbf{x}_t &= \phi^{-(T-t +1)}(\mathbf{h}^{(t)}_T) -\mathbf{b}_{d'} \\
        \Rightarrow \quad
        \mathbf{x}_t 
        &= \mathbf{W}_{d'}^{-1} \Big[
        \phi^{-(T-t +1)}(\mathbf{h}^{(t)}_T) -\mathbf{b}_{d'}
        \Big]
    \end{align}
\end{proof}

\subsection{Notes on Matrix Invertibility}
The above proof is written using $\mathbf{W}_{d'}^{-1}$ for clarity, but exact recovery does not require $\mathbf{W}_{d'}$ to be square or bijective. The essential requirement is that the linear map $\mathbf{W}_{d'} \in \mathbb{R}^{d' \times n}$ be injective.

\paragraph{Continuous inputs.}
For inputs in a continuous domain, exact recovery requires that $\mathbf{W}_{d'}$ have full column rank (i.e., $\mathrm{rank}(\mathbf{W}_{d'}) = n$), which implies $d' \ge n$. Under this condition, a left inverse exists, and $\mathbf{x}_t$ can be recovered uniquely from $\mathbf{W}_{d'} \mathbf{x}_t$. The square full-rank case is a special instance of this more general condition.

If $\mathbf{W}_{d'}$ is not square, recovery can be expressed using the Moore-Penrose pseudoinverse, which yields the exact solution when $\mathbf{W}_{d'}$ has full column rank. If this condition is violated, the mapping is no longer injective and information is irreversibly lost.

\paragraph{Discrete inputs.}
When the inputs lie in a finite set, exact recovery does not require global injectivity of $\mathbf{W}_{d'}$. Instead, it suffices that the mapping be injective over the discrete input set. In practice, one may map the recovered representation (cf.~Eq.~\eqref{eq:rec}) into the embedding space and select the closest valid input under an appropriate metric.

\begin{figure}
    \centering
    \includegraphics[width=\linewidth]{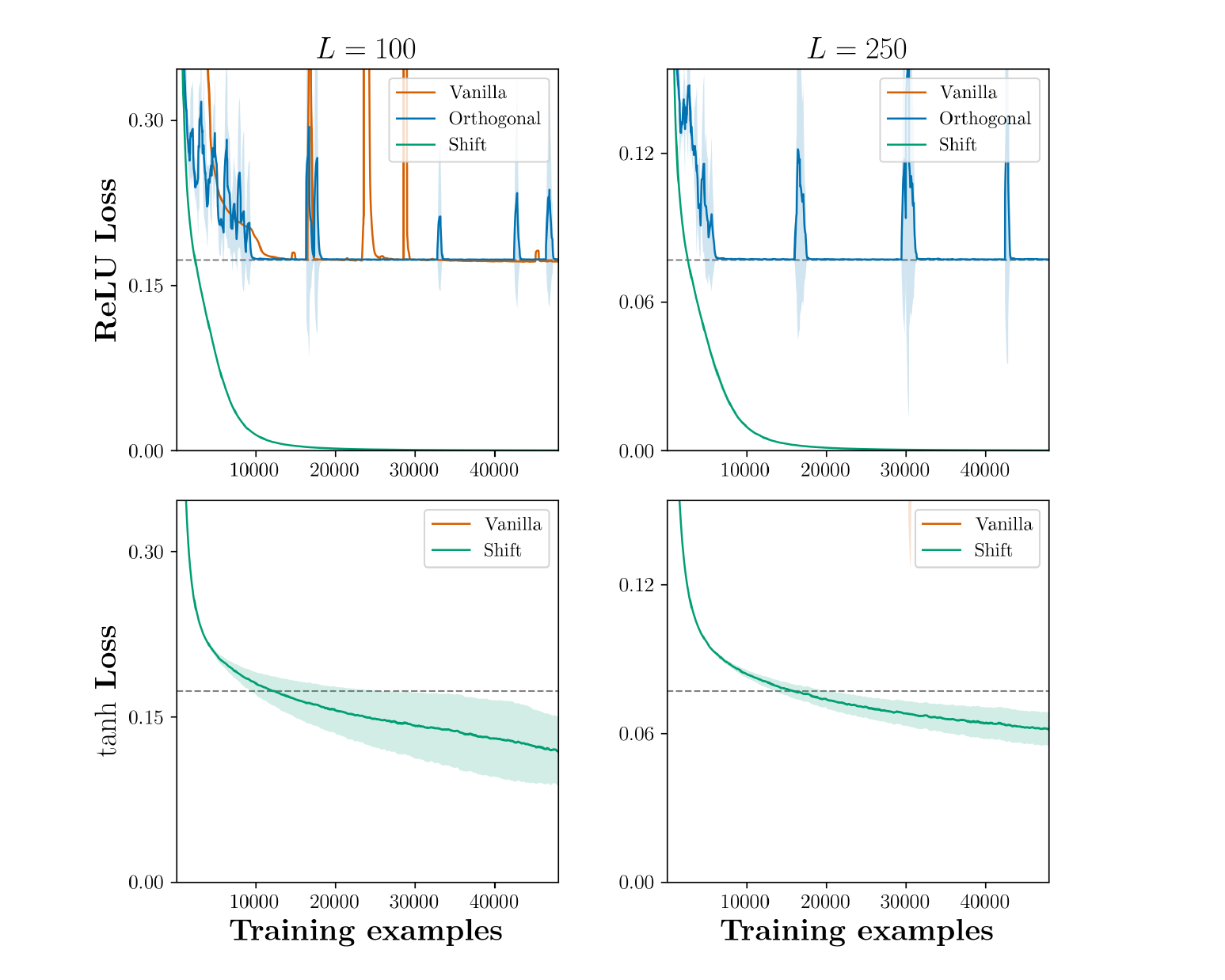}
    \caption{Copy task performance for $L = 100$ and $L = 250$, with ReLU (top row) and $\tanh$ (bottom row): the shift RNN solves both settings with ReLU and slowly surpasses baseline with $\tanh$; the vanilla RNN is stable for ReLU, reaching the baseline.}
    \label{fig:copy_ex}
\end{figure}

\begin{figure}
    \centering
    \includegraphics[width=\linewidth]{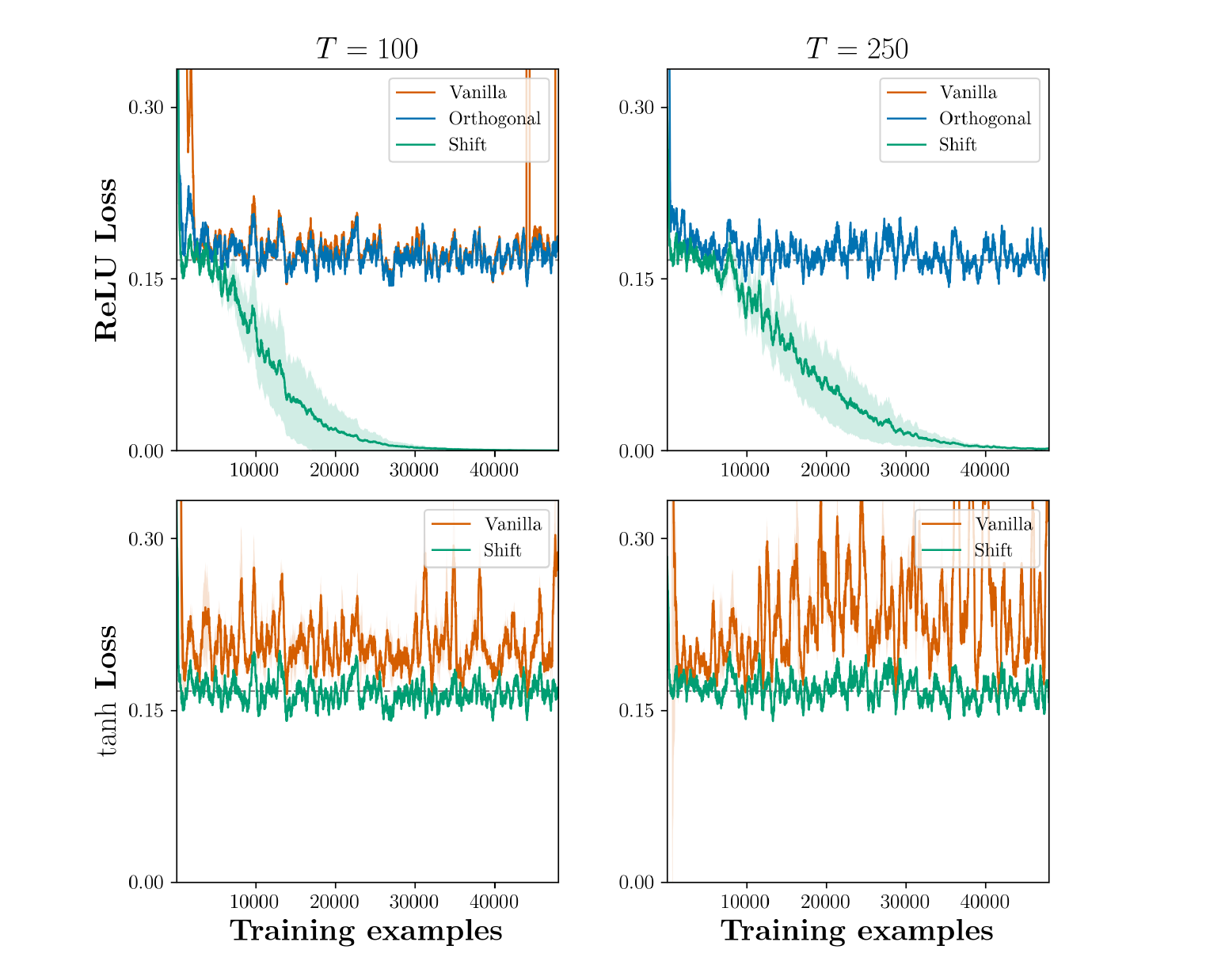}
    \caption{Adding problem performance for $T = 100$ and $T = 250$: with ReLU, the shift RNN solves both settings; with $\tanh$, it does not outperform the baseline.}
    \label{fig:add_ex}
\end{figure}

\begin{figure}
    \centering
    \includegraphics[width=\linewidth]{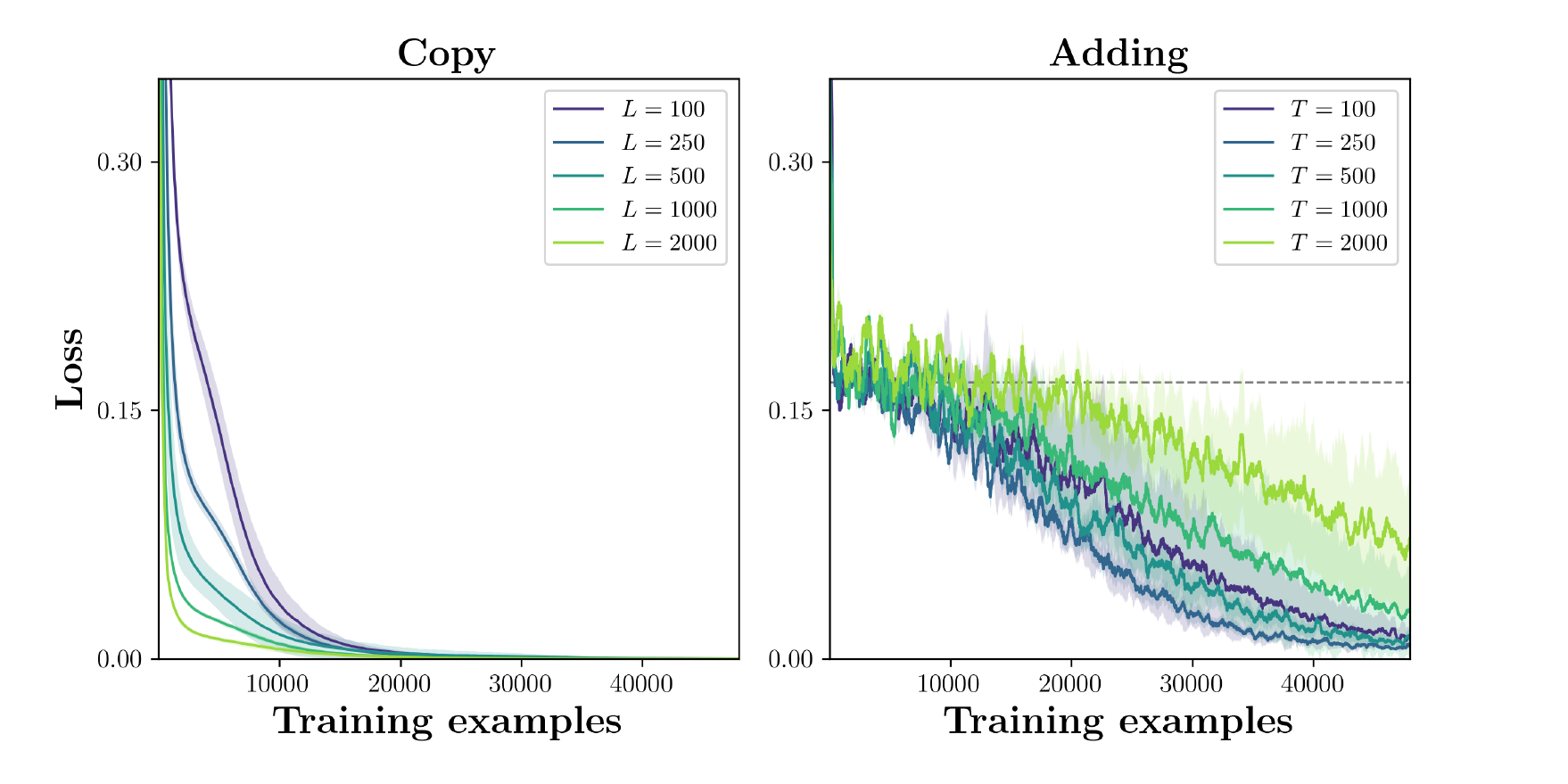}
    \caption{Performance of the shift RNN on the copy task (cross-entropy loss) and adding problem (mean squared error loss) for longer sequence lengths: the model continues to solve both tasks, though learning is slower in the case of the adding problem for $T = 2000$.}
    \label{fig:long}
\end{figure}

\section{Experimental Details}
\label{app:res}

We train all models using Adam with learning rate $0.001$ and batch size $8$. Each task consists of $48{,}000$ training examples. The hidden state dimension is set to $8$ times the sequence length. For the copy task, inputs are represented as one-hot vectors.

We provide additional experimental results to complement those in the main text. First, we extend the main comparison on the copy task and adding problem to additional activation functions, reporting results for ReLU on all, and $\tanh$ on the vanilla and shift RNNs. Secondly, we investigate the shift RNN on longer sequences, evaluating lags and lengths of up to $2000$ to assess its scalability.

We observe a few additional trends across tasks and non-linearities. The vanilla RNN does not exhibit the instability previously seen with ELU when using ReLU, although this may be sensitive to initialisation and run-to-run variability. The orthogonally constrained RNN consistently reaches baseline performance but does not surpass it. The shift RNN fails to solve the adding problem with $\tanh$, but shows an interesting improvement on the copy task, outperforming the baseline; this suggests that numerical precision or optimisation stability may play a role. Finally, the shift RNN continues to outperform baselines as sequence length increases, although the number of training examples required for successful optimisation on the adding task increases gradually with $T$.

\subsection{Computational Resources}

All experiments were run on CPU-only machines. Individual training runs required at most approximately $24$ hours, with memory usage ranging from $2$ GB to $6$ GB for longer shift RNN runs. All experiments were run $3$ times with independent seeds.

\end{document}